\documentclass[journal]{IEEEtran}
\usepackage{amsmath,amsfonts,amssymb,amsthm}
\usepackage{algorithmic}
\usepackage{algorithm}
\usepackage{array}
\usepackage{acronym}
\usepackage{multirow}
\usepackage[caption=false,font=normalsize,labelfont=sf,textfont=sf]{subfig}
\usepackage{textcomp}
\usepackage{stfloats}
\usepackage{url}
\usepackage{verbatim}
\usepackage{graphicx}
\usepackage{cite}
\usepackage{xcolor}
\input{Acronym}
\usepackage{listings}
\usepackage{multirow}
\usepackage{lscape}
\usepackage{hyperref}
\usepackage{circledsteps}
\newcommand\myCircled[2][]{\ifmmode
\Circled[fill color=black,inner color=white,#1]{\textbf{#2}}
\else
\Circled[fill color=black,inner color=white,#1]{\textbf#2}
\fi
}

\definecolor{codegreen}{rgb}{0,0.6,0}
\definecolor{codegray}{rgb}{0.5,0.5,0.5}
\definecolor{codepurple}{rgb}{0.58,0,0.82}
\definecolor{backcolour}{rgb}{0.95,0.95,0.92}
\lstdefinestyle{mystyle}{
    backgroundcolor=\color{backcolour},   
    commentstyle=\color{codegreen},
    keywordstyle=\color{black},
    stringstyle=\color{codepurple},
    basicstyle=\ttfamily\footnotesize,
    breakatwhitespace=false,         
    breaklines=true,                 
    captionpos=b,                    
    keepspaces=true,                             
    showspaces=false,                
    showstringspaces=false,
    showtabs=false,                  
    tabsize=2
}

\newcommand{\cmaroon}[1]{\textcolor{maroon}}

\begin{document}
%\title{Rolling Out Encoder-empowered Multimodal Sensor Transfusion for I2V Beam Prediction}
\title{Multi-Modal Sensor Fusion for Proactive Blockage Prediction in mmWave Vehicular Networks}

\author{~Ahmad M. Nazar,~Abdulkadir~Celik,~Mohamed~Y.~Selim, Asmaa~Abdallah,~Daji~Qiao,~and Ahmed~M.~Eltawil

\thanks{A. M. Nazar, M. Y. Selim, and D. Qiao are with the Department of Electrical and Computer Engineering, Iowa State University of Science and Technology, Ames, IA, 50014, USA.}

\thanks{A. Celik, A. Abdallah, and A. M. Eltawil are with Computer, Electrical, and Mathematical Sciences \& Engineering (CEMSE) Division at King Abdullah University of Science and Technology (KAUST), Thuwal, 23955 KSA. }
}

%The paper headers
%\markboth{Submitted to IEEE Transactions on Mobile Computing}{A. M. Nazar \MakeLowercase{\textit{et al.}}:}

% \IEEEpubid{0000--0000/00\$00.00~\copyright~2025 IEEE}
% % Remember, if you use this you must call \IEEEpubidadjcol in the second
% % column for its text to clear the IEEEpubid mark.

\maketitle

\begin{abstract}
Vehicular communication systems operating in the millimeter wave (mmWave) band are highly susceptible to signal blockage from dynamic obstacles such as vehicles, pedestrians, and infrastructure. To address this challenge, we propose a proactive blockage prediction framework that utilizes multi-modal sensing, including camera, GPS, LiDAR, and radar inputs in an \ac{I2V} setting. This approach uses modality-specific deep learning models to process each sensor stream independently and fuses their outputs using a softmax-weighted ensemble strategy based on validation performance. Our evaluations, for up to 1.5s in advance, show that the camera-only model achieves the best standalone trade-off with an F1-score of 97.1\% and an inference time of 89.8ms. A camera+radar configuration further improves accuracy to 97.2\% F1 at 95.7ms.  Our results display the effectiveness and efficiency of multi-modal sensing for mmWave blockage prediction and provide a pathway for proactive wireless communication in dynamic environments.
\end{abstract}


\section{Introduction}
\Ac{mmWave} frequencies have emerged as key next-generation vehicular communication systems enablers, promising ultra-high data rates and low latency. However, the high-frequency nature of these bands makes them inherently vulnerable to signal blockage from dynamic obstacles such as vehicles and pedestrians. These disruptions are problematic in fast-changing environments, where traditional network handling mechanisms often react too slowly to prevent communication hinderances \cite{fontaine2024towards, zhu2020toward, surveyBeamIEEE}.

Recent research has turned toward generative AI due to its alluring capabilities \cite{letaief2019roadmap}. Methods such as proactive blockage prediction enable preemptive actions such as beam steering, link rerouting, or adjusting \ac{RIS} \cite{abdallah2024multiagentbeamtraining, abdallah2022risaidedmimochannelestimation}. Prediction remains a significant challenge, as it requires understanding the spatial layout and temporal dynamics of complex traffic environments. This challenge motivates the need for real-time sensing and learning frameworks that temporally reason over multiple sensor inputs \cite{dl_survey, letaief2019roadmap}.

Initial efforts to characterize blockage in vehicular settings have relied on analytical models. For example, \cite{blockage1} derives the unconditional blockage probability in highway multi-lane scenarios using parameters such as vehicle dimensions, antenna placement, and traffic density. The work in \cite{blockage2} proposes a radar-based deep learning model that uses object tracking to anticipate future blockages. Similarly, \cite{nazar_sensors} incorporates LiDAR-based user localization into a graph neural network to optimize RIS beamforming. However, their method assumes blockages are already present and focuses on improving communication through relay mechanisms. Additionally, the work in \cite{blockage3} highlights the limitations of beamforming and relay-based solutions, and further emphasizes the importance of proactive sensing in \ac{mmWave} vehicular communication. 

This work proposes a modular, learning-based framework for proactive blockage prediction in \ac{I2V} scenarios. Our architecture integrates time-synchronized data from camera, GPS, LiDAR, and radar sensors co-located with a \ac{BS} at a \ac{RSU}. Each modality is processed independently through a dedicated blockage model, and their predictions are fused using a softmax-weighted ensemble approach based on validation performance. This late fusion strategy enables modularity, interpretability, and real-time feasibility and allows the system to adapt to varying sensing conditions.

The contributions of this work are threefold. First, we formulate the \ac{I2V} blockage prediction problem as a multi-modal prediction task and construct a real-world, temporally aligned sensor processing pipeline using the DeepSense6G dataset~\cite{DeepSense, DeepSense1}. Second, we design and implement modality-specific deep learning models for camera, GPS, LiDAR, and radar data with a confidence-weighted fusion strategy that balances predictive accuracy and computational efficiency. Third, we perform extensive evaluation across multiple sensor configurations. Our findings show that the camera and radar configuration models provide a notable balance between inference latency and predictive performance, making them ideal for real-time vehicular applications.

The remainder of the paper is organized as follows: Section \ref{sec: system} describes the system model and problem formulation. Section \ref{sec: architecture} details the proposed model architecture. Section \ref{sec: eval} presents the models' performance evaluation. Section \ref{sec: conclusion} concludes the paper and discusses directions for future work.

\section{System Model and Problem Formulation}
\label{sec: system}
\begin{figure}
    \centering
    \includegraphics[width=1.0\linewidth]{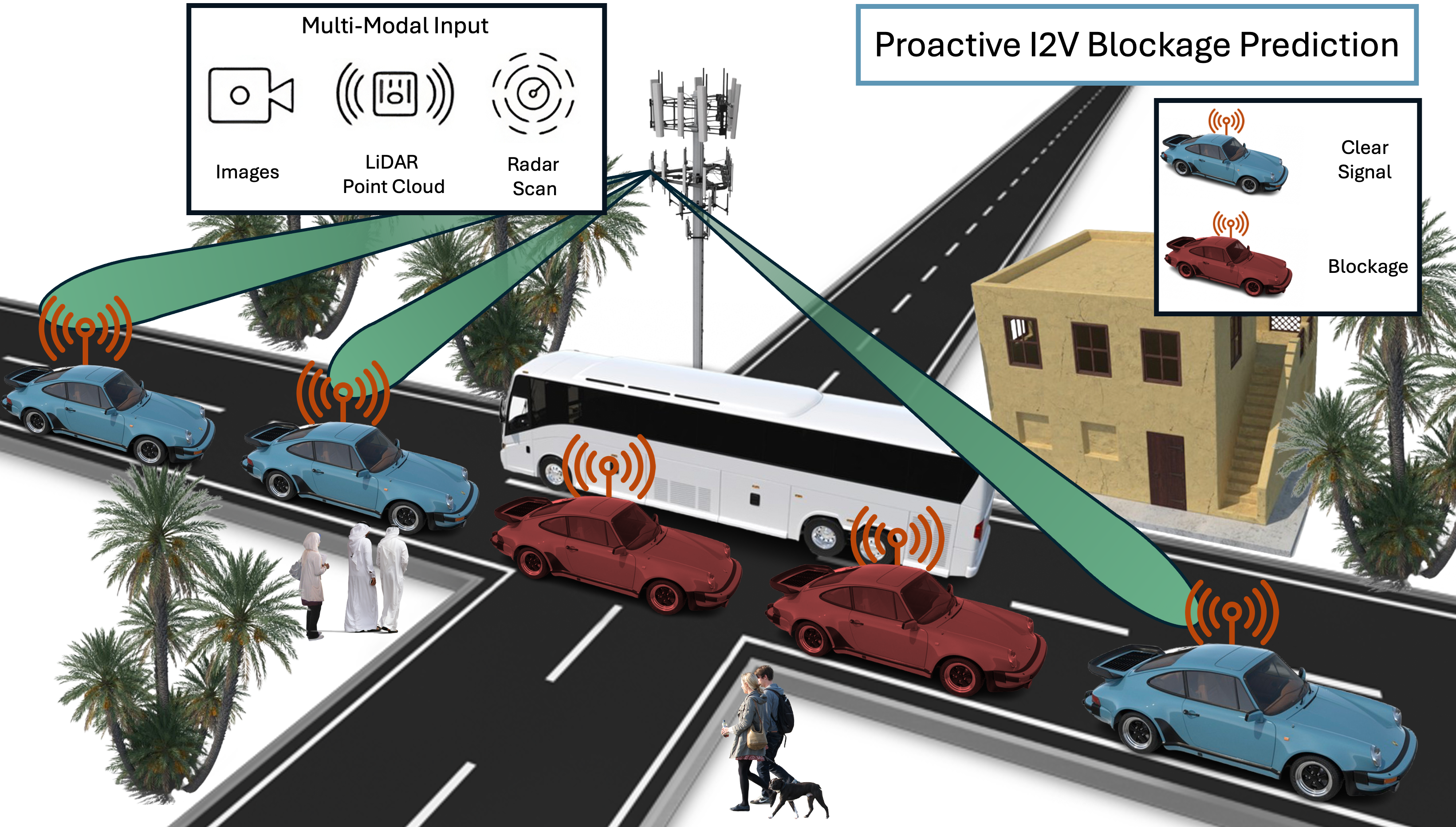}
    \caption{Illustration of the system model.}
    \label{fig:sys_model}
\end{figure}

This work considers a \ac{mmWave} \ac{I2V} communication scenario involving two geo-tagged components, as depicted in Fig.~\ref{fig:sys_model}. The first component is a vehicle equipped with a \ac{ULA} consisting of $M$ antennas. The second component is a fixed \ac{RSU}, which includes a single-antenna \ac{BS} co-located and time-synchronized with a multi-sensor suite comprising a camera, GPS, radar, and LiDAR.

In this setting, the \ac{RSU} periodically captures environmental data using the sensors and transmits it to the blockage model. The model receives fused multi-sensor input from the \ac{RSU} at discrete time steps $t \in \{t_1, t_2, ..., t_n\}$. Each input consists of the latest five sensor observations (i.e., a window of $\Delta T = 1.5$ seconds, and $t=300$ms), denoted as:

\begin{equation}
    \mathcal{X}_t = \left\{ \mathbf{x}_{t-4}, \mathbf{x}_{t-3}, \mathbf{x}_{t-2}, \mathbf{x}_{t-1}, \mathbf{x}_t \right\},
\end{equation}

where $\mathbf{x}_i \in \mathbb{R}^{C \times H \times W}$ represents the features extracted from the multi-modal sensors at time $i$.

The blockage prediction model aims to predict the likelihood of blockage events up to $k$ steps ahead by estimating the blockage probability vector within $k$ denoted as:
\begin{equation}
    \mathbf{p}_t = \left[ p_{t+1}, p_{t+2}, \dots, p_{t+k} \right], \quad \text{where} \quad p_{t+i} \in [0, 1].
\end{equation}

Each element $p_{t+i}$ represents the probability of a blockage occurring at future time $t+i$, given the past sensor window $\mathcal{X}_t$. The model is trained as a binary classifier for each future step, where the labels $y_{t+i} \in \{0, 1\}$ indicate the presence or absence of a blockage event at the corresponding time.

We define the blockage prediction function, $f_{\theta}(\cdot)$, parameterized by $\theta$ as:
\begin{equation}
    \mathbf{p}_t = f_{\theta}(\mathcal{X}_t).
\end{equation}

\section{Proactive Multi-Modal Blockage Prediction}
\label{sec: architecture}
\begin{figure}
    \centering
    \includegraphics[width=1.0\linewidth]{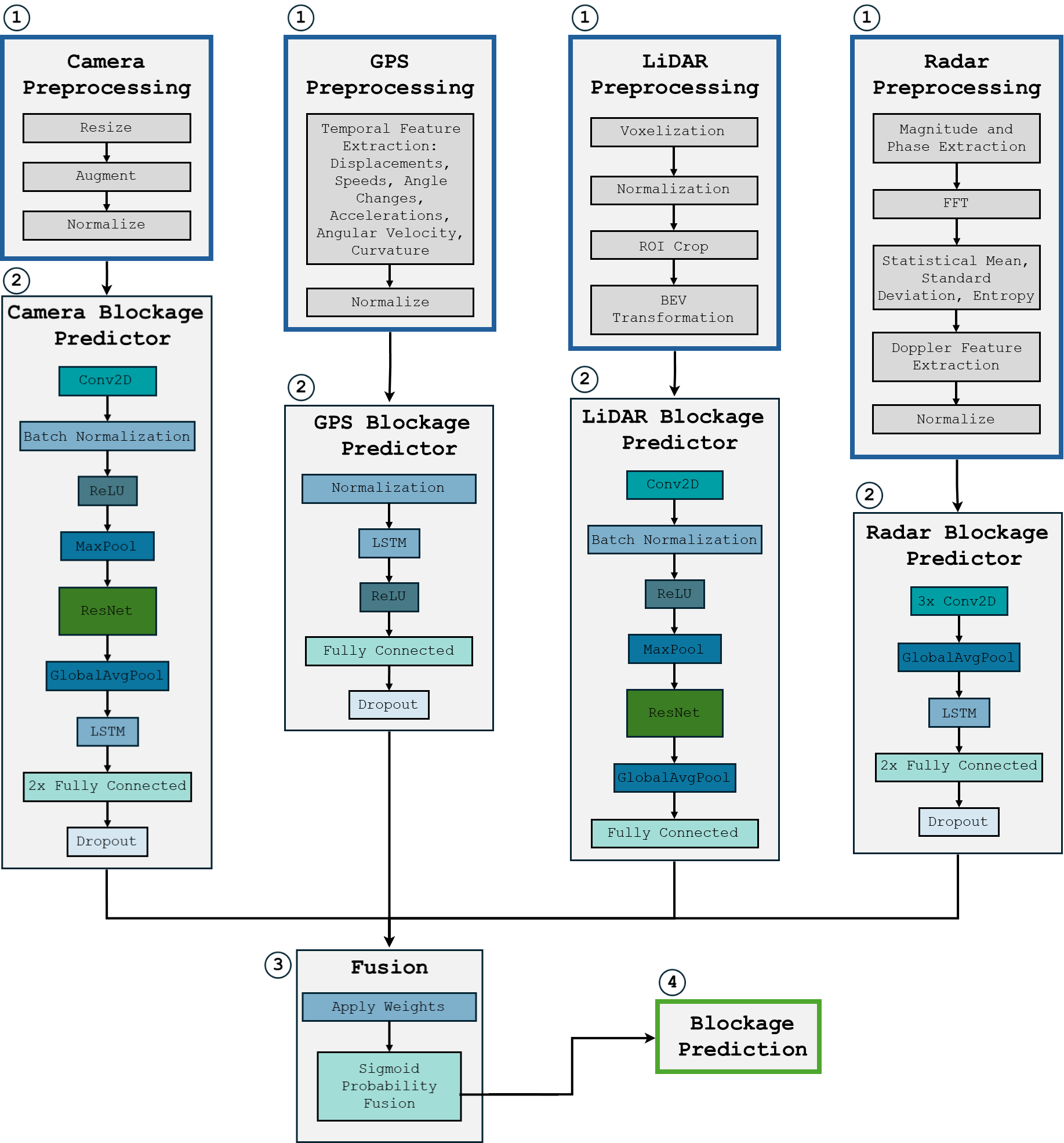}
    \caption{Architecture of the proposed blockage prediction model.}
    \label{fig:model-architecture-blockage-prediction}
\end{figure}

We propose a modular multi-modal architecture for predicting future blockage events in a \ac{mmWave} \ac{I2V} setting through camera, GPS, LiDAR, and radar inputs. As illustrated in Fig.~\ref{fig:model-architecture-blockage-prediction}, the overall pipeline comprises four stages designed to process, infer, and fuse sensory observations temporally.
Stage {\large \textcircled{\small \textbf{1}}\normalsize} performs modality-specific preprocessing to convert raw sensor data into spatially and temporally consistent representations;  
Stage {\large \textcircled{\small \textbf{2}}\normalsize} applies independent deep learning models tailored to each modality to predict blockage probabilities based on past observations;  
Stage {\large \textcircled{\small \textbf{3}}\normalsize} combines the outputs from all modalities using a weighted probability fusion strategy, where each prediction is scaled according to its estimated reliability;  
Finally, Stage {\large \textcircled{\small \textbf{4}}\normalsize} outputs the blockage classification.

This architecture enables each sensor stream to be modeled independently and flexibly fused at the decision level for interpretability, modularity, and resilience to sensor-specific failure or noise. The remainder of this section elaborates on each modality model's architecture and learning strategy, followed by the fusion design that unifies their predictions.

\subsection{Data Preprocessing}
\label{sec:preprocessing}
This work leverages real-world, multi-modal sensor measurements from the DeepSense6G dataset~\cite{DeepSense1,DeepSense2} to construct a robust input representation for blockage prediction. The dataset provides heterogeneous sensor modalities captured in realistic vehicular environments. In particular, we utilize Scenarios 31-34, which include temporally synchronized data streams such as LiDAR point clouds, radar reflections, camera images, and GPS traces from the vehicle and the \ac{RSU}. 

\subsubsection{Data Augmentation}
We apply tailored data augmentation strategies during training to improve generalization across sensor modalities. For images, random horizontal flips, small-angle rotations, and Gaussian blurring are used to simulate variations in perspective and lighting. LiDAR point clouds and projections are augmented through random flips, rotations, and spatial scaling to introduce geometric diversity. For radar data, random Gaussian noise is added to the raw samples to simulate sensor uncertainty and measurement variability.

\subsubsection{Class Imbalance Handling}
The blockage prediction task is inherently imbalanced, with significantly more non-blocked samples (label 0) than blocked ones (label 1). To address this, we apply a loss weighting strategy that penalizes misclassified positive samples more heavily during training. Specifically, we compute a scalar positive class weight as:

\begin{equation}
    w_{\text{pos}} = \alpha \cdot \left( \frac{N_0}{N_1} \right),
\end{equation}

where $N_0$ and $N_1$ denote the number of negative and positive samples, respectively, and $\alpha =1.1$ is a tunable scaling factor used to adjust the degree of compensation.

\subsubsection{Camera Preprocessing} 
 Camera frames are resized to $256 \times 256$ resolution, and pixel values are normalized to ensure consistency in subsequent feature extraction. 

\subsubsection{GPS Preprocessing}
We implement a handcrafted feature extraction pipeline applied over five consecutive GPS readings to extract meaningful motion dynamics from raw GPS trajectories. We compute first- and second-order temporal derivatives from these sequences to characterize vehicle movement. Specifically, we extract displacements, speeds, angular changes, accelerations, angular velocity, and curvature. These features are chosen to capture both linear and rotational motion, indicative of sudden maneuvers or mobility patterns associated with potential blockages. The final feature vector contains 18 elements: six displacement values, four instantaneous speeds, three angular changes, three accelerations, one angular velocity, and one curvature metric. For consistency across the dataset, we apply min-max normalization using a pre-fitted scaler learned from the training set. 

\subsubsection{LiDAR Preprocessing}
A multi-step pipeline is applied to efficiently process the high-density LiDAR point clouds to remove noise, segment objects, and transform raw 3D data into a structured representation suitable for deep learning. 

A voxel-grid downsampling technique reduces the density of LiDAR point clouds while preserving spatial structure \cite{voxel_lidar_ransac}. Outliers in the point cloud are removed using a statistical analysis-based filtering method, where each point’s mean distance to its nearest $k$ neighbors is computed. Points that significantly deviate from the local distribution are discarded based on a threshold derived from a standard deviation factor calculated directly from the point distribution. Following outlier removal, the ground plane is extracted and removed using a \ac{RANSAC} plane-fitting algorithm, which iteratively fits a plane model to the LiDAR points and eliminates those within a fixed distance from the estimated ground surface \cite{voxel_lidar_ransac}.
% These steps ensure that only points corresponding to obstacles and objects of interest remain in the dataset.

To segment individual objects, \ac{DBSCAN} clustering is applied to the filtered point cloud. The clustering parameters, $\epsilon_{\text{lidar}} = 0.75$ and $\texttt{min\_samples}_{\text{lidar}} = 5$, are selected by analyzing a $k$-distance graph, where the optimal $\epsilon_{\text{lidar}}$ corresponds to the maximum curvature point. Clusters not meeting the minimum size criteria are treated as noise and removed.

The processed LiDAR point cloud is then projected into a \ac{BEV} representation, which provides a structured 2D encoding of the scene. The point cloud is mapped onto a fixed grid spanning $X \in [-50, 50]$ m, $Y \in [-50, 50]$ m, and $Z \in [-2.5, 15]$ m, with a discretization resolution of $\Delta r = 0.25$ m per cell. Each grid cell encodes three key features: the maximum $Z$-coordinate of points in the cell to form a height map, the log-normalized count of points per cell to construct a density map, and the variance of height values within the cell to capture surface roughness and structural details. The final BEV image has a resolution of $(H, W) = (700, 1200)$ pixels with three feature channels per frame. Sequential LiDAR frames are stacked along the channel axis, yielding a $(H, W, 15)$ tensor representation representing temporal features.

Filtered point clouds are then normalized such that height values are min-max normalized within the defined $Z$ range, density values are log-scaled to balance sparsity and variance values are standardized based on their maximum observed values. 

\subsubsection{Radar Preprocessing}
We construct a multi-channel feature representation from raw radar data to enhance the discriminative capacity of radar inputs for blockage prediction. Each radar file contains complex-valued tensors from four virtual antenna channels, with a native shape of $(4, 256, 250)$ representing azimuth, range, and Doppler bins, respectively. From these, we extract a comprehensive set of eight feature maps. First, the raw data is decomposed into magnitude and phase components to capture amplitude and phase variation across spatial cells. A 1D Fast Fourier Transform (FFT) is then applied along the Doppler dimension to extract spectral information, particularly the distribution of returned signal energy across velocity bins.

In addition to raw features, we compute several statistical descriptors to encode spatial structure and motion dynamics. These include the mean and standard deviation of magnitude across antenna channels and an entropy map computed from the normalized magnitude, which serves as a proxy for scene complexity or target diversity. Doppler-specific features such as mean velocity and spectral width are derived from the power spectrum to characterize dynamic motion across bins.

All feature maps are normalized to a consistent spatial resolution of $(256 \times 64)$ using trimming or zero-padding to accommodate variable Doppler bin lengths. One-dimensional descriptors are repeated across antenna channels where needed to ensure shape consistency. The final radar input is constructed by stacking all eight features into a tensor of shape $(8, 256, 64)$, capturing a rich blend of spatial, spectral, and kinematic cues essential for robust blockage forecasting.

\subsection{Camera Blockage Prediction}
The camera model is designed to capture spatial and temporal patterns indicative of future blockage events. The spatial feature extraction backbone is a ResNet-18 convolutional neural network pretrained on ImageNet, with the final fully connected classification layer removed. 

To model temporal dynamics, we process the input sequences using an \ac{LSTM}. The \ac{LSTM} allows the model to capture motion cues, object trajectories, and evolving occlusions. For the \ac{LSTM} configuration, the encoded frame embeddings are passed through a single-layer \ac{LSTM} with 128 hidden dimensions, and the final hidden state is used for downstream classification. This layer's output is then passed through a two-layer fully connected classifier with a ReLU activation and dropout regularization ($p = 0.4$), and outputs the blockage probability. 

\subsection{GPS-Based Blockage Prediction Model}
The GPS blockage prediction model is designed to extract mobility features from GPS samples and predict blockages. We extract an 18-dimensional feature vector from the input sequence encoding displacement, velocity, acceleration, angular change, and curvature using a custom feature engineering pipeline. These features are normalized and passed to a temporal encoder composed of a two-layer \ac{LSTM} with a hidden size of 128. The \ac{LSTM}'s final hidden state is processed through a fully connected classifier consisting of a 64-unit hidden layer, ReLU activation, dropout, and outputs the blockage prediction. 
% Despite its simplicity, the GPS model offers a lightweight and interpretable baseline for blockage forecasting, though its coarse spatial resolution and lack of environmental context limit its standalone predictive power.

\subsection{LiDAR Blockage Prediction}
The LiDAR model is built to extract spatial features from a temporally stacked \ac{BEV} representation of LiDAR point clouds. Each input sample comprises five consecutive \ac{BEV} frames containing three spatial channels, resulting in a 15-channel tensor input. We adopt a ResNet-18 convolutional backbone to accommodate this high-dimensional input by modifying the first convolutional layer to accept 15 channels.

The ResNet-18 architecture then processes the stacked LiDAR frames through a hierarchy of residual blocks, progressively extracting deep spatial features relevant to the environment's obstacle presence, object shapes, and occlusion patterns. Unlike the camera and radar models, which incorporate explicit temporal modeling, the temporal information in the LiDAR stream is implicitly learned through the early fusion of the sequential \ac{BEV} frames. The final fully connected layer of the ResNet is modified to output the blockage probability. 

\subsection{Radar Blockage Prediction}
The radar-based model is designed to extract spatial motion signatures and temporal evolution from sequential radar scans. Each radar frame is an 8-channel tensor of size $256 \times 64$, capturing various magnitude, phase, frequency, and Doppler-derived features. These are processed through a stack of three 2D convolutional layers with ReLU activations and batch normalization, progressively reducing spatial resolution while increasing channel depth from 8 to 128.

An adaptive average pooling layer is applied to each radar frame to further condense spatial information. The resulting sequence of feature vectors is then passed through a single-layer \ac{LSTM} with 64 hidden dimensions. The \ac{LSTM}’s final hidden state is input to a fully connected classification head consisting of two linear layers with ReLU activation, dropout regularization ($p = 0.3$), and outputs the blockage probability.

\subsection{Multi-Modal Fusion}
To effectively combine predictions from the LiDAR, radar, and camera models, we adopt a late fusion strategy that avoids the architectural and computational complexity of end-to-end multi-modal networks. Each modality processes its input independently and produces a probability score indicating the likelihood of a future blockage event. These modality-specific predictions—denoted as $P_{\text{LiDAR}}, P_{\text{Radar}},$ and $P_{\text{Camera}}$—are then fused at the decision level. The fusion mechanism is based on a weighted averaging scheme:
\begin{equation}
    P_{\text{fused}} = \sum_{i \in \{\text{LiDAR}, \text{Radar}, \text{Camera}\}} w_i \cdot P_i .
\end{equation}

To determine the weights $w_i$, we adopt a data-driven approach grounded in validation performance. Specifically, each modality’s weight is computed using the softmax over its F1-score obtained from a held-out validation set. Let $\mathbf{s} = [s_{\text{LiDAR}}, s_{\text{Radar}}, s_{\text{Camera}}]$ denote the vector of validation F1-scores. Then, the normalized fusion weights are given by:
\begin{equation}
    \mathbf{w} = \text{softmax}(\mathbf{s}) = \frac{\exp(s_i)}{\sum_j \exp(s_j)}.
\end{equation}

This formulation ensures that higher-performing modalities have more influence in the final prediction while retaining contributions from less dominant modalities. It also avoids the potential bias of manually assigned weights, and adapts to the relative strengths of the models during training.

% The system maintains modularity and interpretability by decoupling the fusion mechanism from joint backpropagation. Individual modality-specific models can be updated or replaced independently without retraining the entire pipeline. This late fusion strategy balances prediction accuracy, model robustness, and real-world feasibility.

\section{Evaluation}
\label{sec: eval}
% Please add the following required packages to your document preamble:
% \usepackage{multirow}
\begin{table*}
\centering
\caption{Blockage prediction model performance summary}
\begin{tabular}{|c|cc|cc|cc|cc|cc|}
\hline
\multirow{2}{*}{\textbf{Modality}} & \multicolumn{2}{c|}{\textbf{$t+1$}} & \multicolumn{2}{c|}{\textbf{$t+2$}} & \multicolumn{2}{c|}{\textbf{$t+3$}} & \multicolumn{2}{c|}{\textbf{$t+4$}} & \multicolumn{2}{c|}{\textbf{$t+5$}} \\ \cline{2-11} 
 & \multicolumn{1}{c|}{\textbf{\begin{tabular}[c]{@{}c@{}}F1 \\ Score\end{tabular}}} & \textbf{AUC-ROC} & \multicolumn{1}{c|}{\textbf{\begin{tabular}[c]{@{}c@{}}F1 \\ Score\end{tabular}}} & \textbf{AUC-ROC} & \multicolumn{1}{c|}{\textbf{\begin{tabular}[c]{@{}c@{}}F1 \\ Score\end{tabular}}} & \textbf{AUC-ROC} & \multicolumn{1}{c|}{\textbf{\begin{tabular}[c]{@{}c@{}}F1 \\ Score\end{tabular}}} & \textbf{AUC-ROC} & \multicolumn{1}{c|}{\textbf{\begin{tabular}[c]{@{}c@{}}F1 \\ Score\end{tabular}}} & \textbf{AUC-ROC} \\ \hline
camera\_radar & \multicolumn{1}{c|}{98.4\%} & 0.988 & \multicolumn{1}{c|}{98.0\%} & 0.985 & \multicolumn{1}{c|}{97.9\%} & 0.982 & \multicolumn{1}{c|}{97.5\%} & 0.971 & \multicolumn{1}{c|}{97.2\%} & 0.968 \\ \hline

camera\_only & \multicolumn{1}{c|}{98.1\%} & 0.983 & \multicolumn{1}{c|}{97.8\%} & 0.981 & \multicolumn{1}{c|}{97.5\%} & 0.970 & \multicolumn{1}{c|}{97.2\%} & 0.969 & \multicolumn{1}{c|}{97.1\%} & 0.963 \\ \hline

camera\_lidar & \multicolumn{1}{c|}{96.2\%} & 0.955 & \multicolumn{1}{c|}{96.1\%} & 0.947 & \multicolumn{1}{c|}{95.7\%} & 0.939 & \multicolumn{1}{c|}{95.5\%} & 0.935 & \multicolumn{1}{c|}{95.5\%} & 0.933 \\ \hline

camera\_gps & \multicolumn{1}{c|}{94.4\%} & 0.931 & \multicolumn{1}{c|}{94.1\%} & 0.930 & \multicolumn{1}{c|}{93.9\%} & 0.927 & \multicolumn{1}{c|}{93.9\%} & 0.927 & \multicolumn{1}{c|}{93.8\%} & 0.924 \\ \hline

camera\_radar\_lidar & \multicolumn{1}{c|}{94.0\%} & 0.928 & \multicolumn{1}{c|}{93.9\%} & 0.926 & \multicolumn{1}{c|}{93.8\%} & 0.925 & \multicolumn{1}{c|}{93.7\%} & 0.922 & \multicolumn{1}{c|}{93.7\%} & 0.920 \\ \hline

radar\_only & \multicolumn{1}{c|}{93.7\%} & 0.922 & \multicolumn{1}{c|}{93.7\%} & 0.921 & \multicolumn{1}{c|}{93.6\%} & 0.919 & \multicolumn{1}{c|}{93.5\%} & 0.916 & \multicolumn{1}{c|}{93.5\%} & 0.915 \\ \hline

camera\_gps\_radar\_lidar & \multicolumn{1}{c|}{93.1\%} & 0.914 & \multicolumn{1}{c|}{93.0\%} & 0.913 & \multicolumn{1}{c|}{92.7\%} & 0.910 & \multicolumn{1}{c|}{92.2\%} & 0.909 & \multicolumn{1}{c|}{92.0\%} & 0.909 \\ \hline

camera\_gps\_radar & \multicolumn{1}{c|}{92.3\%} & 0.911 & \multicolumn{1}{c|}{92.2\%} & 0.907 & \multicolumn{1}{c|}{92.0\%} & 0.904 & \multicolumn{1}{c|}{91.8\%} & 0.902 & \multicolumn{1}{c|}{91.7\%} & 0.898 \\ \hline

radar\_ladar & \multicolumn{1}{c|}{91.8\%} & 0.902 & \multicolumn{1}{c|}{91.6\%} & 0.899 & \multicolumn{1}{c|}{91.5\%} & 0.899 & \multicolumn{1}{c|}{91.4\%} & 0.896 & \multicolumn{1}{c|}{91.3\%} & 0.891 \\ \hline

camera\_gps\_lidar & \multicolumn{1}{c|}{91.2\%} & 0.897 & \multicolumn{1}{c|}{91.1\%} & 0.895 & \multicolumn{1}{c|}{91.1\%} & 0.894 & \multicolumn{1}{c|}{90.8\%} & 0.890 & \multicolumn{1}{c|}{90.8\%} & 0.890 \\ \hline

gps\_lidar\_radar & \multicolumn{1}{c|}{89.9\%} & 0.889 & \multicolumn{1}{c|}{89.8\%} & 0.887 & \multicolumn{1}{c|}{89.5\%} & 0.885 & \multicolumn{1}{c|}{89.4\%} & 0.881 & \multicolumn{1}{c|}{89.0\%} & 0.879 \\ \hline

gps\_radar & \multicolumn{1}{c|}{89.3\%} & 0.880 & \multicolumn{1}{c|}{89.2\%} & 0.879 & \multicolumn{1}{c|}{89.0\%} & 0.877 & \multicolumn{1}{c|}{88.9\%} & 0.875 & \multicolumn{1}{c|}{88.6\%} & 0.873 \\ \hline

lidar\_only & \multicolumn{1}{c|}{87.9\%} & 0.872 & \multicolumn{1}{c|}{87.9\%} & 0.872 & \multicolumn{1}{c|}{87.7\%} & 0.870 & \multicolumn{1}{c|}{87.4\%} & 0.868 & \multicolumn{1}{c|}{87.3\%} & 0.865 \\ \hline

gps\_lidar & \multicolumn{1}{c|}{84.1\%} & 0.855 & \multicolumn{1}{c|}{84.0\%} & 0.852 & \multicolumn{1}{c|}{83.6\%} & 0.849 & \multicolumn{1}{c|}{83.2\%} & 0.847 & \multicolumn{1}{c|}{83.0\%} & 0.831 \\ \hline

gps\_only & \multicolumn{1}{c|}{61.7\%} & 0.603 & \multicolumn{1}{c|}{61.4\%} & 0.600 & \multicolumn{1}{c|}{60.9\%} & 0.599 & \multicolumn{1}{c|}{60.7\%} & 0.592 & \multicolumn{1}{c|}{60.6\%} & 0.588 \\ \hline
\end{tabular}
\label{tab:performance}
\end{table*}

\begin{table}
\centering
\caption{Blockage prediction model timings in milliseconds (ms)}
\begin{tabular}{|c|c|c|c|}
\hline
\textbf{Modality} & \textbf{Preprocessing Time} & \textbf{Inference Time}  \\ \hline
camera\_gps\_radar\_lidar & 56.0 & 137.2    \\ \hline
camera\_radar\_lidar      & 54.3 & 128.9    \\ \hline
camera\_gps\_lidar        & 43.3 & 127.5    \\ \hline
gps\_lidar\_radar         & 51.8 & 125.0    \\ \hline
radar\_lidar              & 50.7 & 124.1    \\ \hline
camera\_lidar             & 42.4 & 121.8    \\ \hline
gps\_lidar                & 38.4 & 121.6    \\ \hline
camera\_gps\_radar        & 18.3 & 121.2    \\ \hline
lidar\_only               & 37.9 & 117.3    \\ \hline
camera\_gps               & 4.96 & 109.4    \\ \hline
gps\_radar                & 13.5 & 96.1     \\ \hline
camera\_radar             & 17.8 & 95.7     \\ \hline
radar\_only               & 12.4 & 92.6     \\ \hline
camera\_only              & 4.11 & 89.8     \\ \hline
gps\_only                 & $<1$ & 37.3     \\ \hline
\end{tabular}
\label{tab:times}
\end{table}

This section evaluates the performance of 15 unordered permutation-based blockage models using F1-score, \ac{AUC-ROC}, and inference time as \acp{KPI}. Each model completes inference within the 300ms interval between consecutive inputs to maintain compatibility with real-time deployment requirements. Table~\ref{tab:performance} summarizes predictive performance, while Table~\ref{tab:times} presents timing performance.

Among all configurations, the camera-only and camera+radar models offer the best trade-offs between accuracy and latency. With inference times of 89.8ms and 95.7ms, respectively, these models are ideally suited for latency-sensitive applications. Focusing on the $t+5$ prediction horizon, camera-based models consistently outperform all other modality combinations. The camera-only model achieves an F1-score of 97.1\% while maintaining the lowest overall latency, demonstrating that visual cues alone provide rich spatiotemporal information sufficient for robust blockage prediction. The camera+radar model further improves performance to 97.2\% F1 through the benefits of radar’s motion and depth perception capabilities without significantly increasing inference latency.

In contrast, including the LiDAR does not provide additional predictive value in high-performing configurations. Although LiDAR contributes detailed 3D structural information, its addition consistently leads to lower F1 scores than camera-only or camera+radar variants. For example, the camera+LiDAR model achieves 95.5\% F1 at $t+5$ while incurring a significantly higher latency of 121.8ms. The marginal structural advantages offered by LiDAR are outweighed by the increased input dimensionality and redundancy, which may lead to overfitting, especially when temporal correlations are weak.

The addition of GPS data similarly offers negligible or even negative value for blockage prediction. Across all GPS-inclusive models, performance degrades relative to models without GPS despite noticeable increases in latency. For instance, the camera+GPS model achieves only 93.8\% F1 at $t+5$, and the full sensor fusion model reaches just 92.0\% F1 which is well below the top performing configurations. These results confirm that GPS-based data lacks the temporal and environmental awareness necessary to predict blockages.

Radar-only models also demonstrate strong performance, achieving 93.5\% F1 at $t+5$, with low inference latency (92.6ms). This result suggests that radar remains a reliable sensing modality. Meanwhile, LiDAR-only models trail behind, reaching only 87.3\% F1, reaffirming their limited standalone effectiveness for the blockage prediction task.

\section{Conclusion}
\label{sec: conclusion}
Our research introduces a proactive blockage prediction framework for \ac{mmWave} \ac{I2V} communication systems through multi-modal sensor fusion. We predict future blockage events up to 1.5 seconds in advance through temporally synchronized multi-modal data with an F1-score of up to 98.4\%.
% We developed deep learning models specific to each modality, tailored to the distinct characteristics of each sensor. Additionally, we implemented a weighted late fusion strategy to combine predictions based on the confidence levels of the models.
Our findings highlight the significance of selecting appropriate modalities in real-time blockage prediction systems, revealing that lightweight, vision-centric configurations are practical and conducive to deployment. Future research aims to investigate online adaptation strategies, quantify uncertainties, and integrate our framework with beam selection to enhance resilience in dynamic vehicular environments.

\bibliographystyle{IEEEtran}
\bibliography{Bibliography/bibliography}

\end{document}